\documentclass[10pt,conference,compsocconf]{IEEEtran}

\usepackage[utf8]{inputenc}
\usepackage[]{algorithm2e}
\usepackage{gensymb}
\usepackage{amssymb}
\usepackage{amsmath}
\usepackage{graphicx}
\usepackage{caption}
\usepackage{subcaption}
\usepackage{wrapfig}
\usepackage{url}
\usepackage{hyperref}
\usepackage{float}
\usepackage{xcolor}
\usepackage{listings}

\begin{document}

\title{A Directional Diffusion Algorithm for Inpainting}

\author{
  Jan Deriu, Rolf Jagerman, Kai-En Tsay\\
  Department of Computer Science, ETH Zürich, Switzerland
}

\maketitle

\begin{abstract}
The problem of inpainting involves reconstructing the missing areas of an image. Inpainting has many applications, such as reconstructing old damaged photographs or removing obfuscations from images.
In this paper we present the directional diffusion algorithm for inpainting. Typical diffusion algorithms are bad at propagating edges from the image into the unknown masked regions. The directional diffusion algorithm improves on the regular diffusion algorithm by reconstructing edges more accurately. It scores better than regular diffusion when reconstructing images that are obfuscated by a text mask. 
\end{abstract}

\section{Introduction}
\label{sec:introduction}

Automatically reconstructing the missing areas of an image is an inference task often referred to as inpainting. It is commonly used to restore old damaged photographs or to remove objects from an image \cite{bertalmio2000image}. A variety of approaches have been proposed in the literature to solve this problem. Many successful algorithms use a diffusion-based approach where local pixel information is propagated into the missing areas of an image \cite{alvarez1992image}. Other approaches involve the reconstruction of the full image based on wavelets or dictionaries \cite{cilhaar, cillearned}.

In this paper we introduce a novel algorithm that solves the inpainting task by using  directional diffusion. This approach is based on the fast diffusion algorithm described by McKenna et al. \cite{richard2001fast}. It attempts to improve on the fast diffusion algorithm by taking into account the directionality of parts of the image to select proper diffusion kernels.

The paper is structured in several sections. First we will describe the methods we use to accomplish the inpainting task in section \ref{sec:methods}. Next we will discuss the results of our methods in section \ref{sec:results} and compare them to several baselines. In section \ref{sec:discussion} the strengths and weaknesses of our novel approach are discussed. Finally the paper is concluded in section \ref{sec:conclusion}.

\section{Methods}
\label{sec:methods}

We present two algorithms that accomplish the inpainting task: regular diffusion and directional diffusion. The first is a naive yet fast approach. The second is an extension of the first, which takes into account the directionality of image patches to perform smarter diffusions. Although it is slower, it typically leads to better results.

\subsection{Regular diffusion}
The idea of diffusion is to fix the known regions of the image and let them diffuse into the unknown regions. This can be done very efficiently by using a convolutional kernel. By iteratively convolving a kernel with the entire image and then restoring the known pixels we can obtain an inpainted image. This process is described in algorithm \ref{alg:diffusion}. The quality of the solution heavily depends on the kernel used. A variety of kernels can be used, two of which are displayed in figure \ref{fig:kernels}. To convolve a kernel with an image we have to refer to pixels outside of the border of the image. We replicate the borders of the image outwards so the kernel can properly refer to these values.

\begin{figure}
\begin{flalign*}
K_{\text{diamond}} &= \begin{bmatrix}0 & 0.25 & 0 \\ 0.25 & 0 & 0.25 \\ 0 & 0.25 & 0\end{bmatrix}\\
K_{\text{diag}} &= \begin{bmatrix}0.38 & 0.04 & 0.04 \\ 0.04 & 0 & 0.04 \\ 0.04 & 0.04 & 0.38\end{bmatrix}
\end{flalign*}
\caption{Kernels used for diffusion.}
\label{fig:kernels}
\end{figure}

\begin{algorithm}
	\KwIn{Image $I$, mask $M$, kernel $K$ and threshold $\epsilon$}
	\KwResult{Reconstructed image $I_{r}$}
	$K \leftarrow \frac{K}{\sum_i \sum_j K_{i,j}}$ (normalize $K$ to preserve energy)\; 
	$I_{prev} \leftarrow 0_{size(I)}$\;
	$I_{r} \leftarrow I$\;
	\While{$\|I_{r} - I_{prev} \|_{F} > \epsilon$}{
		$I_{prev} \leftarrow I_{r}$\;
		$I_{r} \leftarrow \text{convolve}(I_{r}, K)$\;
		$I_{r} \leftarrow I_{r} \circ \mathbf{1}_{M = 0} + I \circ \mathbf{1}_{M \neq 0}$ \;
	}
	\quad
\caption{Diffusion algorithm for inpainting. We denote element-wise multiplication with the $\circ$ operator. The $\mathbf{1}_{M=0}$ function represents a matrix with elements $(i,j)$ set to 1 when $M_{i,j}=0$ and 0 otherwise.}
\label{alg:diffusion}
\end{algorithm}
\begin{figure}
	\centering
	\begin{subfigure}[b]{0.4\textwidth}
		\centering
		\includegraphics[clip, trim=0cm 5.2cm 0cm 4cm, width=0.85\textwidth]{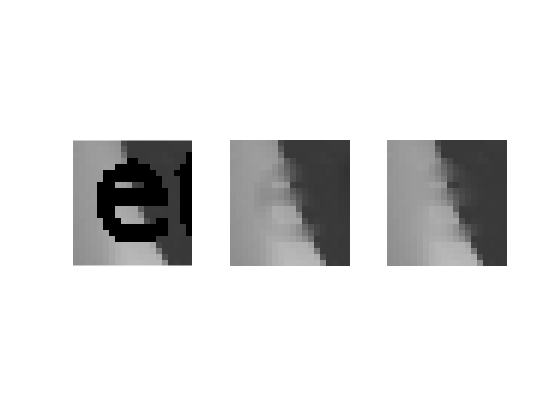}
		\caption{Diamond kernel $K_{\text{diamond}}$}
		\label{fig:stepbystepcross}
	\end{subfigure}
	\begin{subfigure}[b]{0.4\textwidth}
		\centering
		\includegraphics[clip, trim=0cm 5.2cm 0cm 4cm, width=0.85\textwidth]{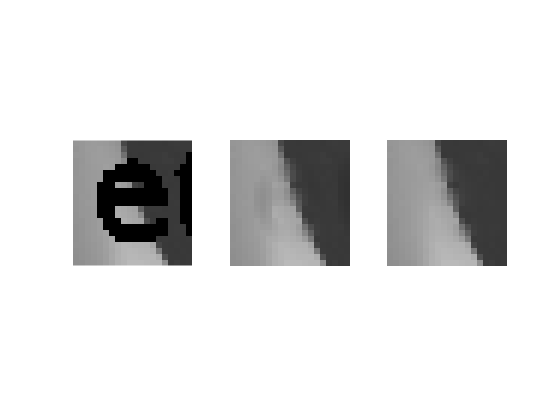}
		\caption{Directional kernel $K_\theta$ for $\theta = 100\degree$.}
		\label{fig:stepbystepdir}
	\end{subfigure}
	\caption{Step-by-step illustration of the diffusion process with different kernels. Each step represent 20 iterations.}
\end{figure}

\subsection{Directional diffusion}

Regular diffusion will present noticeable artifacts. These typically occur at high-contrast edges of the image. Because the diffusion kernel equally weighs pixels from all directions it creates a blur effect and is unable to properly propogate edges into the unknown regions. This effect can be seen in figure \ref{fig:stepbystepcross}. To help resolve this problem we use directional kernels that weigh pixels from certain directions more heavily. We define directionality to be the angle in which the high-contrast edges of an image are directed. An image with mostly horizontal lines will have a directionality of $0$ degrees whereas an image with mostly vertical lines will have a directionality of $90$ degrees. By inferring the correct directionality, the artifacts can be noticeably reduced as seen in figure \ref{fig:stepbystepdir}.

To do this we first apply the regular diffusion algorithm to the image to get an estimate of the original image. Then we divide the image into separate patches. For each patch we infer the directionality using a heuristic. Based on this directionality we construct a directional kernel. Finally we apply this kernel to do the inpainting for that specific patch. We will now explain these steps in more detail.

\begin{figure*}
	\centering
	\begin{subfigure}[b]{0.3\textwidth}
		\centering
		\includegraphics[clip, trim=2cm 0cm 2cm 0cm, width=0.9\textwidth]{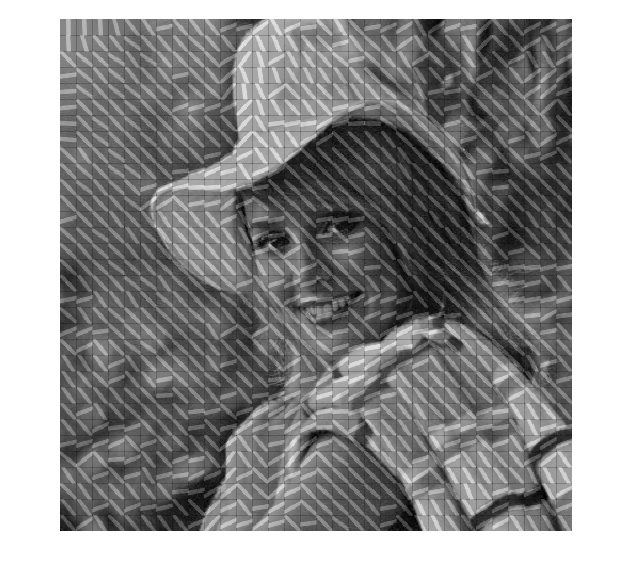}
		\caption{The Elaine image.}
		\label{fig:claudia}
	\end{subfigure}
	\begin{subfigure}[b]{0.3\textwidth}
		\centering
		\includegraphics[clip, trim=2cm 0cm 2cm 0cm, width=0.9\textwidth]{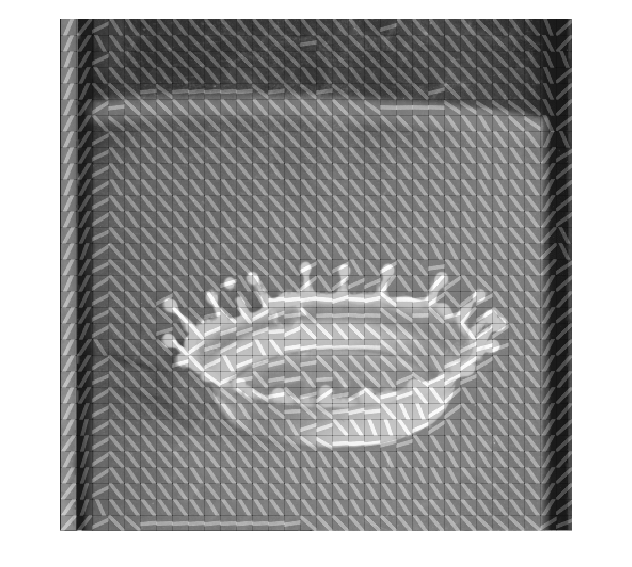}
		\caption{The splash image.}
		\label{fig:spiral}
	\end{subfigure}
	\begin{subfigure}[b]{0.3\textwidth}
		\centering
		\includegraphics[clip, trim=2cm 0cm 2cm 0cm, width=0.9\textwidth]{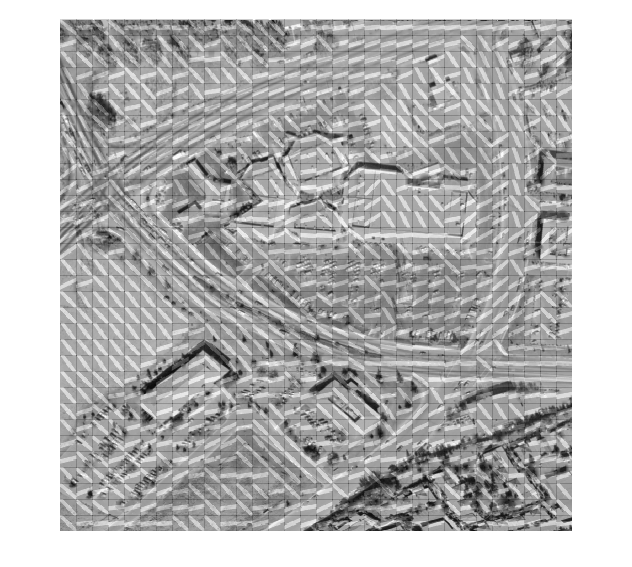}
		\caption{The aerial image.}
		\label{fig:spiral}
	\end{subfigure}
	
	\caption{The directionality of 16x16 pixel patches shown as white lines. Images obtained from the SIPI database \cite{sipiimg}}
	\label{fig:directionality}
\end{figure*}

\subsubsection{Divide image into $n \times n$ patches}
The directionality of the contents of an image is typically a local property. Hence we want to know what the directionality of a small image patch is. An image $I$ can be broken down into patches of size $n \times n$. We define a patch $P$ starting at pixel location $i,j$ as:
\begin{flalign*}
P = \begin{bmatrix}
I_{i,j} & I_{i+1, j} & \hdots & I_{i+n, j}\\
I_{i,j+1} & I_{i+1, j+1} & \hdots & I_{i+n, j+1}\\
\vdots & \vdots & \ddots & \vdots \\
I_{i, j+n} & I_{i+1, j+n} & \hdots & I_{i+n, j+n}
\end{bmatrix}
\end{flalign*}

\subsubsection{Infer directionality of an image patch}
Given an image patch $P$ we wish to compute an estimated angle $\theta$ of the contents of this patch. Intuitively this means for patches with lots of horizontal stripes we would get $\theta \approx 0\degree$ and for patches with lots of vertical stripes we would get $\theta \approx 90\degree$. The first thing we need is a shift operator for matrices. Given a matrix $M \in \mathbb{R}^{n\times m}$ we can compute the shifted matrix $M^{(x,y)}$, which circularly shifts columns towards the right $x$ times and rows towards the bottom $y$ times:
\begin{flalign*}
M^{(x, y)}_{i, j} &= M_{(i + x) \text{ mod } n, (j + y)\text{ mod } m}
\end{flalign*}
Using this shift operator, we compute two basic metrics on the image patch:
\begin{flalign*}
v &= \sum_{i, j}  | P_{i,j} - P_{i,j}^{(1, 0)} | \\
h &= \sum_{i, j} | P_{i,j} - P_{i,j}^{(0, 1)} | 
\end{flalign*}
The first metric $v$ checks how much each pixel differs from the one left of it and provides large values for images with a lot of vertical stripes. The second metric $h$ does the same but checks how much each pixel differs from the one below it. This gives large values for images with a lot of horizontal stripes.

By comparing $h$ and $v$ we can determine if the contents of the image patch will have mostly vertical lines ($v > h$), mostly horizontal lines ($v < h$) or diagonal lines ($v \approx h$). We can turn this into an angle, by computing the fraction of $h$ to $h + v$.  We add $1$ to both sides to prevent division by zero:
\begin{flalign*}
\theta_1 &= 90 \frac{h + 1}{h + v + 1}
\end{flalign*}

This approach only gives an angle from 0$\degree$ to 90$\degree$. It cannot infer if the diagonal lines on the image patch are left-slanted or right-slanted since these will have $h \approx v$. To help resolve this we propose a heuristic for the diagonal slant by computing $d$, which checks how much each pixel differs from the one diagonally right-below it. This gives large values for images with right-slanted diagonal lines but small values for images with left-slanted diagonal lines. We wish to normalize this value by dividing it by $v+h$, so it ranges from 0 to 1. We add 1 to both sides to prevent division by zero:
\begin{flalign*}
d &= \frac{1 + \sum_{i, j} | P_{i,j} - P_{i,j}^{(1, 1)} |}{1 + v + h}
\end{flalign*}

If this value $d$ is close to 1, we can assume the image is right-slanted. If it is close to 0, we can assume the image is left-slanted. A value close to 0.5 indicates that the image patch has both left and right slanted lines. We can now compute the angle $\theta$ with the additional parameter $d$ using the following heuristic:
\begin{flalign*}
\theta &=  - 90 + \begin{cases}
90d+\theta_1 & \text{if } d > 0.6 \\
90-\theta_1       & \text{if } d \leq 0.6
\end{cases}
\end{flalign*}

In figure \ref{fig:directionality} we show the angle $\theta$, which has been computed on image patches of size $16 \times 16$, as white lines. The heuristic works well on image patches where the contents is simple and consists of mostly straight lines. When the content of an image patch gets more complex, inferring the angle becomes more difficult and the heuristic fails.

\subsubsection{Constructing a directional kernel}
After computing the angle $\theta$ for an image patch, we wish to construct a directional kernel $K_\theta$, which puts more weight on pixels that align with the angle $\theta$. We start with a diagonal kernel $K_{\text{diag}}$ (see figure \ref{fig:kernels}). This kernel is converted to a $3\times 3$ image which is rotated by $\theta+45$ degrees (because the initial kernel was diagonal, we add 45 degrees). It is then cropped to obtain a new $3\times 3$ image representing the rotated kernel. During the rotation process, bicubic interpolation is used to smooth the intermediate values \cite{keys1981cubic}. The resulting image is converted back to a $3 \times 3$ matrix representing the kernel $K_\theta$.

\subsubsection{Per-patch diffusion using $K_\theta$}
Now that we have a kernel $K_\theta$ for each image patch $P$ we can run the diffusion algorithm (see algorithm \ref{alg:diffusion}) on each of the patches. The reconstructed image is created by putting all the patches back together after running the diffusion algorithm on them.

\section{Results}
\label{sec:results}

We test our methods on seventeen different $512\times 512$ grayscale images using six different algorithms:
\begin{enumerate}
	\item Sparse-coding with a DCT dictionary \cite{cildct}.
	\item Sparse-coding with a Haar wavelet \cite{cilhaar}.
	\item Singular Value Decomposition \cite{cilsvd}.
	\item Regular diffusion with a $K_{\text{diamond}}$ kernel.
	\item Directional diffusion with patches of size $16 \times 16$.
	\item Directional diffusion with patches of size $32 \times 32$.
\end{enumerate}

 To test the performance of the algorithm, we consider two types of masks:
\begin{enumerate}
	\item Nine different masks of randomly missing pixels distributed uniformly. These range from 10\% to 90\% missing pixels. This type of mask has many small missing regions.
	\item A mask represented by a sample of text. This type of mask has many medium-sized missing regions.
\end{enumerate}

We have set up the experiments as follows: We take the original image $I$ and a mask $M$. We construct a damaged image $I_{\text{damaged}}$ based on $I$ but with the pixels indicated by the mask $M$ set to the fixed value 0. We run the relevant inpainting algorithms by providing it with both the damaged image and the mask. The algorithm will return a recovered image $I^{\text{rec}}$, which we can compare to the original image $I$.

We compare the algorithms based on two criteria: the mean-squared error and the runtime of the algorithm. The mean-squared error is computed as the mean of the square of the difference in all pixel intensity values of the original image and the reconstructed image:
\begin{flalign*}
\text{MSE}(I, I^{\text{rec}}) = \frac{1}{512 \cdot 512} \sum_{i,j} (I_{i,j} - I^{\text{rec}}_{i,j})^2
\end{flalign*}

The mean squared error of the algorithms for the randomly generated masks are displayed in figure \ref{fig:err_random} and the runtime is displayed in figure \ref{fig:runtime}. The mean squared error and runtime of the algorithms for the text mask is given in table \ref{tbl:err_text}.

\begin{table}
	\centering
	\begin{tabular}{|l|c|c|}
		\hline
		\textbf{Algorithm} & \textbf{MSE} & \textbf{Runtime} \\ \hline \hline
		Directional Diffusion ($16 \times 16$) & $\mathbf{0.00055} \pm 0.00051$ & $8.7 \pm 2.13$ \\ \hline
		Directional Diffusion ($32 \times 32$) & $0.00057 \pm 0.00053$ & $2.4 \pm 0.04$ \\ \hline
		Diffusion ($K_{\text{diamond}}$) & $0.00061 \pm 0.00057$ & $\mathbf{0.5} \pm 0.06$ \\ \hline
		Sparse-coding (DCT) & $0.0015 \pm 0.0012$ & $12.8 \pm 4.67$ \\ \hline
		Sparse-coding (Haar wavelet) & $0.0024 \pm 0.0021$ & $13.0 \pm 3.72$ \\ \hline
		Singular Value Decomposition & $0.0019 \pm 0.0018$ & $0.7 \pm 0.09$ \\ \hline
	\end{tabular}
	\caption{Mean squared error and runtime (in seconds) across different algorithms for the text mask. The best result is highlighted in bold.}
	\label{tbl:err_text}
\end{table}


\section{Discussion}
\label{sec:discussion}

\begin{figure*}
	\centering
	\begin{subfigure}[b]{0.49\textwidth}
		\centering
		\includegraphics[clip, trim=2cm 7cm 2cm 6cm, width=0.85\textwidth]{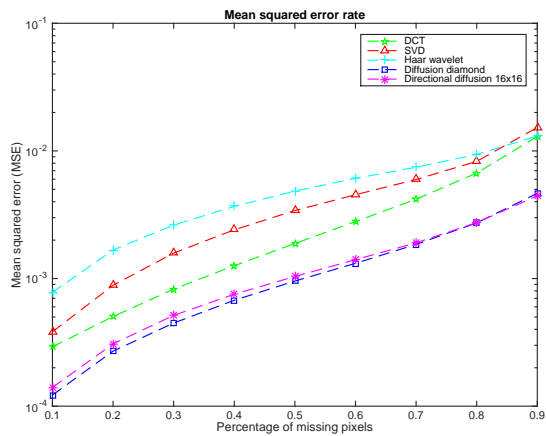}
		\caption{Mean squared error in log scale.}
		\label{fig:err_random}
	\end{subfigure}
	\begin{subfigure}[b]{0.49\textwidth}
		\centering
		\includegraphics[clip, trim=2cm 7cm 2cm 6cm, width=0.85\textwidth]{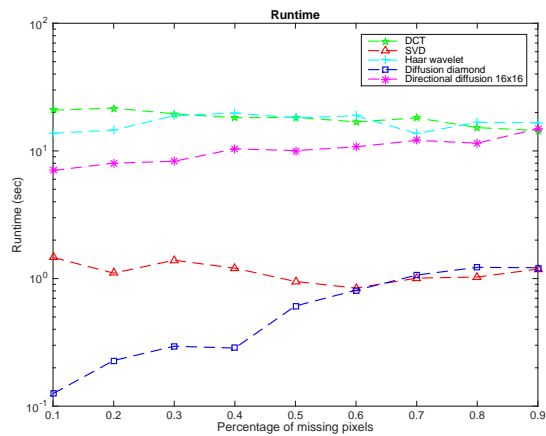}
		\caption{Runtime in log scale.}
		\label{fig:runtime}
	\end{subfigure}
	
	\caption{Mean squared error and runtime comparison of different algorithms. To improve readability we only consider the directional diffusion algorithm for patches of size $16 \times 16$.}
	\label{fig:rmd_results}
\end{figure*}

Both the regular diffusion and the directional diffusion algorithm show promising results. From figure \ref{fig:err_random} we see that the regular diffusion algorithm with the $K_{\text{diamond}}$ kernel works best on a mask with randomly missing pixels. This can be explained by the fact that this algorithm diffuses nearby pixels into the missing regions. A mask with randomly missing pixels will, on average, have at least some pixels in the direct or near neighborhood of a pixel that we try to inpaint.

Table \ref{tbl:err_text} shows the mean squared error of the algorithms on a structured mask, namely a piece of text. The regular diffusion algorithm does not perform as well in this setting. This is because it relies on nearby known pixels which are less common in masks with medium-sized areas of missing pixels. As a consequence, the high-contrasting edges of the underlying image are not properly extended into the unknown regions. The directional diffusion algorithm helps resolve this by aligning the kernel  $K_{\theta}$ with the directionality of the image patches. Because of this the directional diffusion algorithm scores best.


\section{Conclusion}
\label{sec:conclusion}

In this paper we presented the directional diffusion algorithm for the inpainting problem. It is an extension to the regular diffusion algorithm. Directional diffusion outperforms regular diffusion when applied to text masks.

The main drawback of the directional diffusion algorithm is its runtime. However, due to the nature of the algorithm, it is very easy to parallelize. By applying the patch-dependent kernels $K_\theta$ on all patches simultaneously we can achieve great speed-ups.

Additionally, the heuristic used to infer the directionality $\theta$ is not perfect and can be improved. One could use the sobel operator to compute gradients of an image patch \cite{sobel2014history}. Based on these gradients it might be possible to get a more robust estimate of the directionality $\theta$ of the image patches.

\bibliographystyle{IEEEtran}
\bibliography{directional-diffusion}

\end{document}